\documentclass[11pt,a4paper]{article}
\usepackage[hyperref]{naaclhlt2018}
\usepackage{times}
\usepackage{latexsym}
\usepackage{array}
\usepackage{multirow}
\usepackage{amsmath}
\usepackage{booktabs}

\usepackage{url}

\usepackage{graphicx}

\aclfinalcopy 

\title{Fast and Scalable Expansion of Natural Language Understanding Functionality for Intelligent Agents}

\author{Anuj Goyal, Angeliki Metallinou, Spyros Matsoukas \\
 Amazon Alexa\\
  {\tt \{anujgoya, ametalli, matsouka\}@amazon.com} \\ 
  }

\date{}

\begin{document}
\maketitle
\begin{abstract}
Fast expansion of natural language functionality of intelligent virtual agents is critical for achieving engaging and informative interactions. However, developing accurate models for new natural language domains is a time and data intensive process. We propose efficient deep neural network architectures that maximally re-use available resources through transfer learning. Our methods are applied for expanding the understanding capabilities of a popular commercial agent and are evaluated on hundreds of new domains, designed by internal or external developers. We demonstrate that our proposed methods significantly increase accuracy in low resource settings and enable rapid development of accurate models with less data.
\end{abstract}

\section{Introduction}
\label{sec:introduction}

Voice powered artificial agents have become widespread among consumer devices, with agents like Amazon Alexa, Google Now and Apple Siri being popular and widely used. Their success relies not only on accurately recognizing user requests, but also on continuously expanding the range of requests that they can understand. An ever growing set of functionalities is critical for creating an agent that is engaging, useful and human-like.

This presents significant scalability challenges regarding rapidly developing the models at the heart of the natural language understanding (NLU) engines of such agents. Building accurate models for new functionality typically requires collection and manual annotation of new data resources, an expensive and lengthy process, often requiring highly skilled teams. In addition, data collected from real user interactions is very valuable for developing accurate models but without an accurate model already in place, the agent will not enjoy widespread use thereby hindering collection of high quality data. 

Presented with this challenge, our goal is to speed up the natural language expansion process for Amazon Alexa, a popular commercial artificial agent, through methods that maximize re-usability of resources across areas of functionality. Each area of Alexa's functionality, e.g., Music, Calendar, is called a \textit{domain}. Our focus is to a) increase accuracy of low resource domains b) rapidly build new domains such that the functionality can be made available to Alexa's users as soon as possible, and thus start benefiting from user interaction data. To achieve this, we adapt recent ideas at the intersection of deep learning and transfer learning that enable us to leverage available user interaction data from other areas of functionality. 



To summarize our contributions, we describe data efficient deep learning architectures for NLU that facilitate knowledge transfer from similar tasks. We evaluate our methods at a much larger scale than related transfer learning work in NLU, for fast and scalable expansion of hundreds of new natural language domains of Amazon Alexa, a commercial artificial agent. We show that our methods achieve significant performance gains in low resource settings and enable building accurate functionality faster during early stages of model development by reducing reliance on large annotated datasets.


\section{Related Work}
\label{sec:related}

Deep learning models, including Long-Short term memory networks (LSTM) ~\cite{gers1999learning}, are state of the art for many natural language processing tasks (NLP), such as sequence labeling \cite{chung_rnn}, named entity recognition (NER) ~\cite{chiu2015named} and part of speech (POS) tagging ~\cite{huang2015bidirectional}. Multitask learning is also widely applied in NLP, where a network is jointly trained for multiple related tasks. Multitask architectures have been succesfully applied for joint learning of NER, POS, chunking and supertagging tasks, as in ~\cite{collobert2011natural, collobert_weston_08, sogaard2016deep}. 


Similarly, transfer learning addresses the transfer of knowledge from data-rich source tasks to under-resourced target tasks. Neural transfer learning has been successfully applied in computer vision tasks where lower layers of a network learn generic features that are transferred well to different tasks ~\cite{zeiler2014visualizing, krizhevsky2012imagenet}. Such methods led to impressive results for image classification and object detection ~\cite{donahue2014decaf, sharif2014cnn, girshick2014rich}  In NLP, transferring neural features across tasks with disparate label spaces is relatively less common. In ~\cite{transferable2016}, authors conclude that network transferability depends on the semantic relatedness of the source and target tasks. In cross-language transfer learning, \cite{buys_botha_2016} use weak supervision to project morphology tags to a common label set, while \cite{kim_emnlp_2017} transfer lower layer representations across languages for POS tagging. Other related work addresses transfer learning where source and target share the same label space, while feature and label distributions differ, including deep learning methods \cite{glorot2011domain, Kim2017AdversarialAO}, and earlier domain adaptation methods such as EasyAdapt~\cite{daume2007frustratingly}, instance weighting~\cite{jiang2007instance} and structural correspondence learning ~\cite{blitzer2006domain}. 






Fast functionality expansion is critical in industry settings. Related work has focused on scalability and ability to learn from few resources when developing a new domain, and includes zero-shot learning~\cite{chen2016zero, ferreira2015zero}, domain attention \cite{kim2017domain}, and scalable, modular classifiers \cite{opendomainLiTur}. There is a multitude of commercial tools for developers to build their own custom natural language applications, including Amazon Alexa ASK \cite{kumar_ask}, DialogFlow by Google~\cite{dialogflow} and LUIS by Microsoft~\cite{luis}. Along these lines, we propose scalable methods that can be applied for rapid development of hundreds of low resource domains across disparate label spaces.



\section{NLU Functionality Expansion}
\label{sec:expansion}

We focus on Amazon Alexa, an intelligent conversational agent that interacts with the user through voice commands and is able to process requests on a range of natural language domains, e.g., playing music, asking for weather information and editing a calendar. In addition to this \textit{built-in} functionality that is designed and built by internal developers, the Alexa Skills Kit (ASK) \cite{kumar_ask} enables external developers to build their own \textit{custom} functionality which they can share with other users, effectively allowing for unlimited new capabilities. Below, we describe the development process and challenges associated with natural language domain expansion.


For each new domain, the internal or external developers define a set of \textit{intents} and \textit{slots} for the target functionality. Intents correspond to user intention, e.g., `FindRecipeIntent', and slots correspond to domain-specific entities of interest e.g.,`FoodItem'. Developers also define a set of commonly used utterances that cover the core use cases of the functionality, e.g., `find a recipe for chicken'. We call those \textit{core utterances}. In addition, developers need to create \textit{gazetteers} for their domain, which are lists of slot values. For example, a gazetteer for `FoodItem' will contain different food names like `chicken'. We have developed infrastructure to allow internal and external teams to define their domain, and create or expand linguistic resources such as core utterances and gazetteers. We have also built tools that enable extracting \textit{carrier phrases} from the example utterances by abstracting the utterance slot values, such as `find a recipe for \{FoodItem\}'. The collection of carrier phrases and gazetteers for a domain is called a \textit{grammar}. Grammars can be sampled to generate synthetic data for model training. For example, we can generate the utterance `find a recipe for pasta' if the latter dish is contained in the `FoodItem' gazetteer.

Next, developers enrich the linguistic resources available for a new domain, to cover more linguistic variations for intents and slots. This includes creating bootstrap data for model development, including collecting utterances that cover the new functionality, manually writing variations of example utterances, and expanding the gazetteer values. In general, this is a time and data intensive process. External developers can also continuously enrich the data they provide for their custom domain. However, external developers typically lack the time, resources or expertise to provide rich datasets, therefore in practice custom domains are significantly under-resourced compared to built-in domains.


Once the new domain model is bootstrapped using the collected datasets, it becomes part of Alexa's natural language functionality and is available for user interactions. The data from such user interactions can be sampled and annotated in order to provide additional targeted training data for improving the accuracy of the domain. A good bootstrap model accuracy will lead to higher user engagement with the new functionality and hence to a larger opportunity to learn from user interaction data. 


Considering these challenges, our goal is to reduce our reliance on large annotated datasets for a new domain by re-using resources from existing domains. Specifically, we aim to achieve higher model accuracy in low resource settings and accelerate new domain development by building good quality bootstrap models faster.

\vspace{-0.1cm}
\section{Methodology}
\label{sec:methods}

In this section, we describe transfer learning methods for efficient data re-use. Transfer learning refers to transferring the knowledge gained while performing a task in a \textit{source} domain $D_s$ to benefit a related task in a \textit{target} domain $D_t$. Typically, we have a large dataset for $D_s$, while $D_t$ is an under-resourced new task. Here, the target domain is the new built-in or custom domain, while the source domain contains functionality that we have released, for which we have large amounts of data. The tasks of interest in both $D_s$ and $D_t$ are the same, namely slot tagging and intent classification. However $D_s$ and $D_t$ have different label spaces $Y_s$ and $Y_t$, because a new domain will contain new intent and slot labels compared to previously released domains.

\subsection{DNN-based natural language engine}
\label{sec:dnn_engine}

\begin{figure}
  \includegraphics[width=\linewidth]{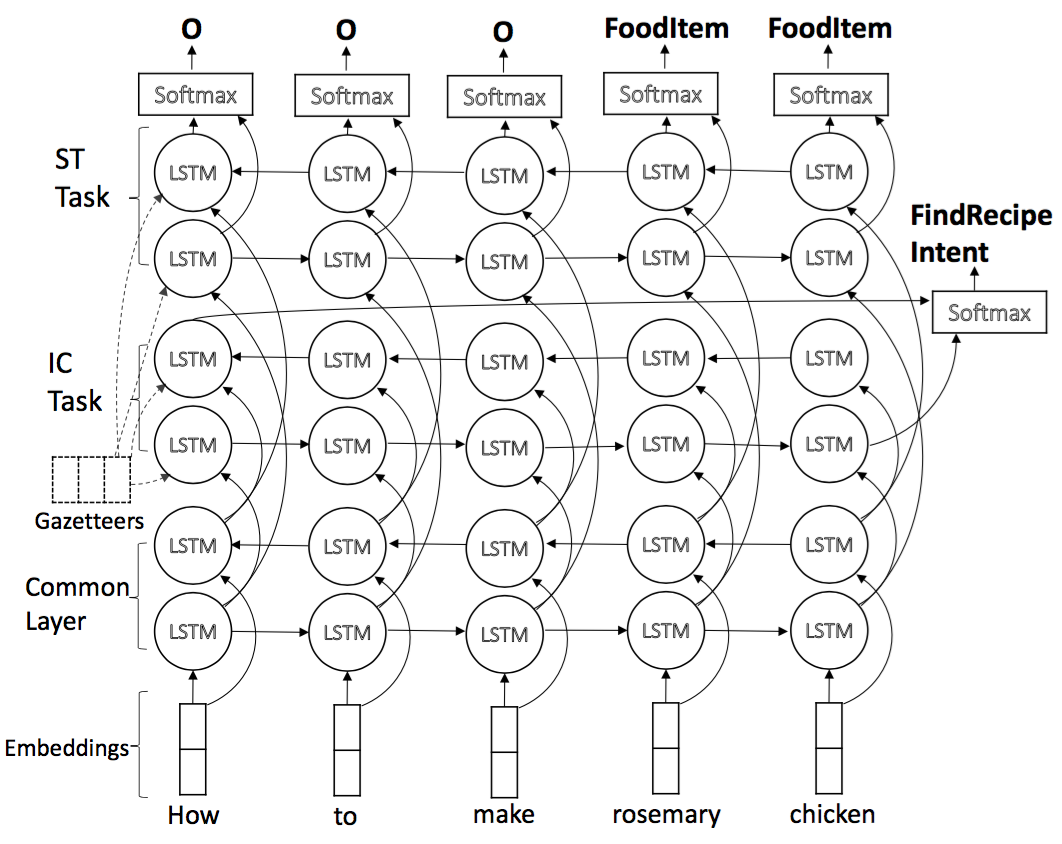}
  \caption{Multitask stacked bi-LSTM architecture for ST and IC, with a shared bottom layer, two separate top layers for ST and IC. Gazetteer features can be added as optional input to the ST and IC layers during the fine-tuning stage. (see also Sec. \ref{sec:transfer_engine})}
  \label{fig:network}
  \vspace{-0.2cm}
\end{figure}

We first present our NLU system where we perform slot tagging (ST) and intent classification (IC) for a given input user utterance. We are inspired by the neural architecture of \cite{sogaard2016deep}, where a multi-task learning architecture is used with deep bi-directional Recurrent Neural Networks (RNNs). Supervision for the different tasks happens at different layers. Our neural network contains three layers of bi-directional Long Short Term Memory networks (LSTMs) ~\cite{graves2005framewise,hochreiter1997long}. The two top layers are optimized separately for the ST and IC tasks, while the common bottom layer is optimized for both tasks, as shown in Figure \ref{fig:network}.

Specifically let $r^{c}_t$ denote the common representation computed by the bottommost bi-LSTM for each word input at time $t$. The ST forward LSTM layer learns a representation $r^{ST, f}_t = \phi ( r^c_t, r^{ST}_{t-1})$, where $\phi$ denotes the LSTM operation. The IC forward LSTM layer learns $r^{IC, f}_t = \phi ( r^c_t, r^{IC}_{t-1})$. Similarly, the backward LSTM layers learn $r^{ST, b}_t$ and $r^{IC, b}_t$. To obtain the slot tagging decision, we feed the ST bi-LSTM layer's output per step into a softmax, and produce a slot label at each time step (e.g., at each input word). For the intent decision, we concatenate the last time step from the forward LSTM with the first  step of the backward LSTM, and feed it into a softmax for classification:
\begin{gather*}
r^{slot}_t = r^{ST,f}_{t} \oplus r^{ST,b}_{t}, r^{intent} = r^{IC,f}_{T} \oplus r^{IC,b}_{1} \\
\hat{S}_t=softmax( W_s r^{slot}_t + b_s) \\
\hat{I}=softmax( W_I r^{intent} + b_I)
\end{gather*}
where $\oplus$ denotes concatenation. $W_s, W_I, b_s, b_I$ are the weights and biases for the slot and intent softmax layers respectively. $\hat{S}_t$ is the predicted slot tag per time step (per input word), and $\hat{I}$ is the predicted intent label for the sentence.


The overall objective function for the multitask network combines the IC and ST objectives. Therefore we jointly learn a shared representation $r^c_t$ that leverages the correlations between the related IC and ST tasks, and shares beneficial knowledge across tasks. Empirically, we have observed that this multitask architecture achieves better results than separately training intent and slot models, with the added advantage of having a single model, and a smaller total parameter size.

In our setup, each input word is embedded into a $300$-dimensional embedding, where the embeddings are estimated from our data. We also use pre-trained word embeddings as a separate input, that allows incorporating unsupervised word information from much larger corpora (FastText~ \cite{bojanowski2016enriching}). We encode slot spans using the IOB tagging scheme \cite{W95-0107}. When we have available \textit{gazetteers} relevant to the ST task, we use gazetteer features as an additional input. Such features are binary indicators of the presence of an n-gram in a gazetteer, and are common for ST tasks \cite{radford2015,nadeau07}.


\subsection{Transfer learning for the DNN engine}
\label{sec:transfer_engine}

Typically, a new domain $D_t$ contains little available data for training the multitask DNN architecture of Sec \ref{sec:dnn_engine}. We propose to leverage existing data from mature released domains (source $D_s$) to build generic models, which are then adapted to the new tasks (target $D_t$).



We train our DNN engine using labeled data from $D_s$ in a supervised way. The source slot tags space $Y^{slot}_s$ and intent label space $Y^{intent}_s$ contain labels from previously released slots and intents respectively. We refer to this stage as \textit{pre-training}, where the stacked layers in the network learn to generate features which are useful for the ST and IC tasks of $D_s$. Our hypothesis is that such features will also be useful for $D_t$. After pre-training is complete, we replace the top-most affine transform and softmax layers for IC and ST with layer dimensions that correspond to the target label space for intents and slots respectively, i.e., $Y^{intent}_t$ and $Y^{slot}_t$. The network is then trained again using the available target labeled data for IC and ST. We refer to this stage as \textit{fine-tuning} of the DNN parameters for adapting to $D_t$.

A network can be pre-trained on large datasets from $D_s$ and later fine tuned separately for many low resource new domains $D_t$. In some cases, when developing a new domain $D_t$, new domain-specific information becomes available, such as domain gazetteers (which were not available at pre-training). To incorporate this information during fine-tuning, we add gazetteer features as an extra input to the two top-most ST and IC layers, as shown in Figure \ref{fig:network}. We found that adding new features during fine-tuning significantly changes the upper layer distributions. Therefore, in such cases, it is better to train the ST and IC layers from scratch and only transfer and fine-tune weights from the common representation $r_c$ of the bottom layer. However, when no gazetteers are available, it is beneficial to pre-train all stacked Bi-LSTM layers (common, IC and ST), except from the task-specific affine transform leading to the softmax.


\subsection{Baseline natural language engine}
\label{sec:baseline_engine}

While DNNs are strong models for both ST and IC, they typically need large amounts of training data. As we focus on under-resourced functionality, we examine an alternative baseline that relies on simpler models; namely a Maximum Entropy (MaxEnt) \cite{berger1996maximum} model for intent classification and a Conditional Random Field (CRF) \cite{lafferty2001conditional} model for slot tagging. MaxEnt models are regularized log-linear models that have been shown to be effective for text classification tasks \cite{berger1996maximum}. Similarly, CRFs have been popular tagging models in the NLP literature \cite{nadeau07} prior to the recent growth in deep learning. In our experience, these models require less data to train well and represent strong baselines for low resource classification and tagging tasks.


\vspace{-0.1cm}
\section{Experiments and Results}
\label{sec:results}


We evaluate the transfer learning methods of Section \ref{sec:transfer_engine} for both custom and built-in domains, and compare with baselines that do not benefit from knowledge transfer (Sections \ref{sec:dnn_engine}, \ref{sec:baseline_engine}). We experiment with around 200 developer defined custom domains, whose statistics are presented in Table \ref{tab:stats}. Looking at the median numbers, which are less influenced by a few large custom domains compared to mean values, we note that typically developers provide just a few tens of example phrases and few tens of values per gazetteer (slot gazetteer size). Therefore, most custom domains are significantly under-resourced. We also select three new built-in domains, and evaluate them at various early stages of domain development. Here, we assume that variable amounts of training data gradually become available, including bootstrap and user interaction data. 

We pre-train DNN models using millions of annotated utterances from existing mature built-in domains. Each annotated utterance has an associated domain label, which we use to make sure that the pre-training data does not contain utterances labeled as any of the custom or built-in target domains. After excluding the target domains, the pre-training data is randomly selected from a variety of mature Alexa domains covering hundreds of intents and slots across a wide range of natural language functionality. For all experiments, we use L1 and L2 to regularize our DNN, CRF and MaxEnt models, while DNNs are additionally regularized with dropout. 

The test sets contain user data, annotated for each custom or built-in domain. For custom domains, test set size is a few hundred utterances per domain, while for built-in domains it is a few thousand utterances per domain. Our metrics include standard F1 scores for the SC and IC tasks, and a sentence error rate (SER) defined as the ratio of utterances with at least one IC or ST error over all utterances. The latter metric combines IC and ST errors per utterance and reflects how many utterances we could not understand correctly. 


\begin{table}[h!]
\vspace{-0.1cm}
	\centering
	\fontsize{9}{10}\selectfont
    \begin{tabular}{c|c|c}
		\toprule
      Data type & Mean & Median\\
		\midrule
      number of intents & 8.02 & 3\\
      number of slots & 2.07 & 1\\
      slot gazetteer size & 441.35 & 11\\
      number of example phrases & 268.11 & 42\\
		\bottomrule
	\end{tabular}
	\caption{Statistics of data for around 200 developer defined custom domains}
	\label{tab:stats}
\end{table}


\vspace{-0.1cm}
\subsection{Results for custom developer domains}
\label{sec:results_custom}

\begin{table*}[t]
\vspace{-0.2cm}
	\centering
	\fontsize{9}{10}\selectfont
	\begin{tabular}{c|cc|cc|cc}
		\toprule
\multirow{2}{*}{Approach} & \multicolumn{2}{c|}{$F1_{Intent}$} &  \multicolumn{2}{c|}{$F1_{Slot}$} &  \multicolumn{2}{c}{$SER$}\\ 		
& Mean & Median & Mean & Median & Mean & Median\\
		\midrule
Baseline CRF/MaxEnt  & 94.6 & 96.6 & 80.0 & 91.5 & 14.5 & 9.2\\
		\midrule
Baseline DNN  & 91.9 & 95.9 & 85.1 & 92.9 & 14.7 & 9.2\\
		\midrule
Proposed Pretrained DNN  * & \bfseries 95.2 & \bfseries 97.2 & \bfseries 88.6 & \bfseries 93.0 & \bfseries 13.1 & \bfseries 7.9\\
		\bottomrule
	\end{tabular}
	\caption{Results for around 200 custom developer domains. For F1, higher values are better, while for SER lower values are better. * denotes statistically significant SER difference compared to both baselines.}
	\label{tb:skillresults}
	\vspace{-0.4cm}
\end{table*}

For the custom domain experiments, we focus on a low resource experimental setup, where we assume that our only target training data is the data provided by the external developer. We report results for around 200 custom domains, which is a subset of all domains we support. We compare the proposed transfer learning method, denoted as \textit{DNN Pretrained}, with the two baseline methods described in sections \ref{sec:dnn_engine} and \ref{sec:baseline_engine}, denoted as \textit{DNN Baseline} and \textit{CRF/MaxEnt Baseline}, respectively. For training the baselines, we use the available data provided by the developer for each domain, e.g., example phrases and gazetteers. From these resources, we create grammars and we sample them to generate 50K training utterances per domain, using the process described in Section \ref{sec:expansion}. This training data size was selected empirically based on baseline model accuracy. The generated utterances may contain repetitions for domains where the external developer provided a small amount of example phrases and few slot values per gazetteer. For the proposed method, we pre-train a DNN model on 4 million utterances and fine tune it per domain using the 50K grammar utterances of that domain and any available gazetteer information (for extracting gazetteer features). In Table \ref{tb:skillresults}, we show the mean and median across custom domains for $F1_{slot}$, $F1_{intent}$ and $SER$.

Table \ref{tb:skillresults} shows that the CRF and MaxEnt models present a strong baseline and generally outperform the DNN model without pretraining, which has a larger number of parameters. This suggests that the baseline DNN models (without pretraining) cannot be trained robustly without large available training data. The proposed pre-trained DNN significantly outperforms both baselines across all metrics (paired t-test, $p < .01$). Median $SER$ reduces by around 14\% relative when we use transfer learning compared to both baselines. We are able to harness the knowledge obtained from data of multiple mature source domains $D_s$ and transfer it to our under-resourced target domains $D_t$, across disparate label spaces.  

To investigate the effect of semantic similarity across source and target domains we selected a subset of 30 custom domains with high semantic  similarity with the source tasks. Semantic similarity was computed by comparing the sentence representations computed by the common bi-LSTM layer across source and target sentences, and selecting target custom domains with sentences close to at least one of the source tasks. For these 30 domains, we observed higher gains of around 19\% relative median SER reduction. This corroborates observations of \cite{transferable2016}, that neural feature transferability for NLP depends on the semantic similarity between source and target. In our low resource tasks, we see a benefit from transfer learning and this benefit increases as we select more semantically similar data.


Our approach is scalable and is does not rely on manual domain-specific annotations, besides developer provided data. Also, pretrained DNN models are about five times faster to train during the fine-tuning stage, compared to training the model from scratch for each custom domain, which speeds up model turn-around time.

\vspace{-0.1cm}
\subsection{Results for built-in domains}
\label{sec:results_builtin}

We evaluate our methods on three new built-in domains referred here as domain A (5 intents, 36 slot types), domain B (2 intents, 17 slot types) and domain C (22 intents, 43 slot types).  Table \ref{tab:domains_results} shows results for domains A, B and C across experimental early stages of domain development, where different data types and amounts of data per data type gradually become available. Core data refers to core example utterances, bootstrap data refers to domain data collection and generation of synthetic (grammar) utterances, and user data refers to user interactions with our agent. As described in Section \ref{sec:expansion}, the collection and annotation of these data sources is a lengthy process. Here we evaluate whether we can accelerate the development process by achieving accuracy gains in early, low resource stages, and bootstrap a model faster.

For each data setting and size, we compare our proposed pretrained DNN models with the baseline CRF/MaxEnt baseline, which is the better performing baseline of Section \ref{sec:results_custom}. Results for the non pre-trained DNN baseline are similar, and omitted for lack of space. Our proposed DNN models are pre-trained on 4 million data from mature domains and then fine tuned on the available target data. The baseline CRF/MaxEnt models are trained on the available target data. Note that the datasets of Table \ref{tab:domains_results} represent early stages of model development and do not reflect final training size or model performance. The types of target data slightly differ across domains according to domain development characteristics. For example, for domain B there was very small amount of core data available and it was combined with the bootstrap data for experiments.

\begin{table}[t]
	\centering
	\fontsize{8}{10}\selectfont
	\begin{tabular}{ccc|c|c|c}
		\toprule
		Train Set & Size & Method & $F1_{intent}$ & $F1_{slot}$ & $SER$ \\
		\midrule
		\midrule
		\multicolumn{6}{c}{Domain A (5 intents, 36 slots)}\\
		Core* & \multirow{2}{*}{500} & Baseline & 85.0 & 63.9 & 51.9 \\
		data  &  & Proposed & 86.6 & 66.6 & 48.2 \\
		\midrule
		Bootstrap & \multirow{2}{*}{18K} & Baseline & 86.1 & 72.8 & 49.6 \\
		data* &  & Proposed & 86.9 & 73.8 & 47.0 \\
		\midrule
		Core + & \multirow{2}{*}{3.5K} & Baseline & 90.4 & 74.3 & 40.5 \\
		user data*  &  & Proposed & 90.1 & 75.8 & 37.9 \\
		\midrule
		Core + & \multirow{3}{*}{43K} & Baseline & 92.1 & 80.6 & 33.4 \\
		bootstrap +  &  & Proposed & 91.9 & 80.8 & 32.8 \\
		user data  &  &  &  &  &  \\
		\midrule
		\midrule
		\multicolumn{6}{c}{Domain B (2 intents, 17 slots)}\\
		Bootstrap & \multirow{2}{*}{2K} & Baseline & 97.0 & 94.7 & 10.1 \\
		data* &  & Proposed & 97.8 & 95.3 & 6.3 \\
		\midrule
		User data & \multirow{2}{*}{2.5K} & Baseline & 97.0 & 94.7 & 8.2 \\
		 &  & Proposed & 97.1 & 96.4 & 7.1 \\
		\midrule
		Bootstrap + & \multirow{2}{*}{52K} & Baseline & 96.7 & 95.2 & 8.2 \\
	    user data* &  & Proposed & 97.0 & 96.6 & 6.4 \\
		\midrule
		\midrule
		\multicolumn{6}{c}{Domain C (22 intents, 43 slots)}\\
		Core* & \multirow{2}{*}{300} & Baseline & 77.9 & 47.8 & 64.2 \\
		data &  & Proposed & 85.6 & 46.6 & 51.8 \\
		\midrule
		Bootstrap & \multirow{2}{*}{26K} & Baseline & 46.1 & 65.8 & 64.0 \\
		data* &  & Proposed & 49.1 & 68.9 & 62.8 \\
		\midrule
		Core + & \multirow{3}{*}{126K} & Baseline & 92.3 & 78.3 & 28.1 \\
		bootstrap. +  &  & Proposed & 92.7 & 72.7 & 31.9 \\
		user data*  &  &  &  &  &  \\
		\bottomrule
	\end{tabular}
	\caption{Results on domains A, B and C for the proposed pretrained DNN method and the baseline CRF/MaxEnt method during experimental early stages of domain development. * denotes statistically significant SER difference between proposed and baseline}
	\label{tab:domains_results}
\end{table}

Overall, we notice that our proposed DNN pretraining method improves performance over the CRF/MaxEnt baseline, for almost all data settings. As we would expect, we see the largest gains for the most low resource data settings. For example, for domain A, we observe a 7\% and 5\% relative SER improvement on core and bootstrap data settings respectively. The performance gain we obtain on those early stages of development brings us closer to our goal of rapidly bootstrapping models with less data.  From domains A and C, we also notice that we achieve the highest performance in settings that leverage user data, which highlights the importance of such data. Note that the drop in $F_{intent}$ for domain C between core and bootstrap data is because the available bootstrap data did not contain data for all of the 22 intents of domain C. Finally, we notice that the gain from transfer learning diminishes in some larger data settings, and we may see degradation (domain C, 126K data setting). We hypothesize that as larger training data becomes available it may be better to not pre-train or pre-train with source data that are semantically similar to the target. We will investigate this as part of future work.



\section{Conclusions and Future Work}
\label{sec:conclusions}

We have described the process and challenges associated with large scale natural language functionality expansion for built-in and custom domains for Amazon Alexa, a popular commercial intelligent agent. To address scalability and data collection bottlenecks, we have proposed data efficient deep learning architectures that benefit from transfer learning from resource-rich functionality domains. Our models are pre-trained on existing resources and then adapted to hundreds of new, low resource tasks, allowing for rapid and accurate expansion of NLU functionality. In the future, we plan to explore unsupervised methods for transfer learning and the effect of semantic similarity between source and target tasks.



\bibliography{naaclhlt2018}
\bibliographystyle{acl_natbib}

\end{document}